\documentclass{article}

 \usepackage[preprint,nonatbib]{neurips_2026}

\usepackage[utf8]{inputenc} 
\usepackage[T1]{fontenc}    
\usepackage{hyperref}       
\usepackage{url}            
\usepackage{booktabs}       
\usepackage{amsfonts}       
\usepackage{nicefrac}       
\usepackage{microtype}      
\usepackage{xcolor}         
\usepackage{hyperref}
\usepackage{amsmath}
\usepackage{graphicx}
\usepackage{microtype}      
\usepackage{xcolor}         
\usepackage{multirow}
\usepackage{pifont}
\usepackage{colortbl}
\usepackage{verbatim}
\usepackage{ulem}
\usepackage{float}
\usepackage{subcaption}

\newcommand{\xmark}{\ding{55}}

\usepackage{amsmath}

\hypersetup{
    colorlinks=true,
    citecolor=blue,   
    linkcolor=blue,   
    urlcolor=blue     
}
\title{TRIMMER: A New Paradigm for Video Summarization through Self-Supervised
Reinforcement Learning}

\author{
Pritam Mishra$^{1}$ \quad
Coloma Ballester$^{2}$ \quad
Dimosthenis Karatzas$^{2}$ \\
\vspace{0.5em}
$^{1}$ Universitat Pompeu Fabra, Barcelona, Spain \\
$^{2}$ Universitat Autònoma de Barcelona, Barcelona, Spain
}

\begin{document}

\maketitle

\begin{abstract}
The rapid growth of video content across domains such as surveillance, education, and social media has made efficient content understanding increasingly critical. Video summarization addresses this challenge by generating concise yet semantically meaningful representations, but existing approaches often rely on expensive manual annotations, struggle to generalize across domains, and incur significant computational costs due to complex architectures. Moreover, unsupervised and weakly supervised methods typically underperform compared to supervised counterparts in capturing long-range temporal dependencies and semantic structure. In this work, we propose TRIMMER (Temporal Relative Information Maximization for Multi-objective Efficient Reinforcement), a novel self-supervised reinforcement learning framework for video summarization. TRIMMER operates in two stages: it first learns robust representations via self-supervised learning and then performs spatio-temporal decision making through reinforcement learning guided by information-theoretic reward functions. Unlike prior approaches that rely on similarity-based objectives, our method introduces entropy-based metrics to capture higher-order temporal dynamics and semantic diversity, while computing rewards directly over selected frame indices to improve computational efficiency. Extensive experiments on standard benchmarks demonstrate that TRIMMER achieves state-of-the-art performance among unsupervised and self-supervised methods, while remaining competitive with leading supervised approaches, highlighting its effectiveness for scalable and generalizable video summarization.

\end{abstract}

\section{Introduction}\label{sec:intro}
The exponential growth of video content across domains such as surveillance, education, and social media has created a pressing need for more efficient methods of content consumption. Video summarization addresses this challenge by producing temporally compressed yet semantically rich representations, thereby reducing the cognitive burden of prolonged viewing and enabling users to comprehend essential content efficiently. Such summaries support the efficiency of a wide range of subsequent tasks such as content discovery, semantic indexing, video retrieval, surveillance monitoring, and user-specific recommendations. Thus, video summarization is not merely a convenience but a necessity for managing video data in a scalable and efficient manner.

Despite recent advances in deep learning, current state-of-the-art video summarization methods continue to face several persistent challenges. Firstly, temporal redundancy and visual similarity among video frames make it difficult to distinguish between semantically meaningful and trivial content without high-level scene understanding, often resulting in summaries that lack temporal coherence and are not readily interpretable or suitable for human consumption. Secondly, existing methods often struggle with generalization across domains, exhibiting performance degradation when applied to video genres different from those seen during training. Thirdly, many existing approaches are either computationally intensive \cite{sspvs,vjmht} or require extensive manual annotations\cite{vasnet,a2summ,rrstg}, limiting their scalability in real-world applications. Although unsupervised \cite{zhou2018deep,gansum,gansum_aae,csnet} and weakly-supervised models \cite{sspvs,ma} have been proposed to mitigate annotation costs, they frequently underperform compared to supervised counterparts \cite{csta,msva,dmasum}, particularly in capturing complex temporal dependencies and long-range semantic cues. To address these challenges, we propose a novel self-supervised reinforcement learning framework, \textbf{TRIMMER} (Temporal Relative Information Maximization for Multi-objective Efficient Reinforcement), which is both scalable and efficient. TRIMMER eliminates the reliance on dataset-specific manual annotations while achieving state-of-the-art performance among unsupervised and self-supervised approaches, and delivering competitive results relative to leading supervised methods on standard video summarization benchmarks.

Our contributions in this paper can be summarized as follows.
\begin{itemize}
    \item This paper introduces a two-stage framework for video summarization: the first stage focuses on self-supervised representation learning, while the second stage leverages the acquired representations to guide spatio-temporal decision making through reinforcement learning.
    \item Our stage-two design incorporates two reward functions centered on diversity and representativeness, with a fundamental methodological distinction from prior work \cite{zhou2018deep}: we utilize entropy-based metrics derived from information theory, in contrast to dot product formulations that capture only first-order linear dependencies.
    \item While our approach draws inspiration from the success of TRIM \cite{trim} in the first stage, it diverges significantly in the second stage not only by our reinforcement learning strategy but also by the fact that our reward functions are computed exclusively over the selected indices rather than across all image frames, enhancing efficiency during training.
\end{itemize}

\section{Related work}\label{sec:relwork}
Video summarization techniques have been extensively studied and applied to generate concise representations of videos, often in the form of storyboards comprising key frames or compilations of the most salient video segments \cite{ego_sum,video_attn,gansum,dpplstm,lan2023collaborative,ego_sum,mem_video,ramos2022text,ramos2020straight}.  Despite significant progress, many existing methods struggle to preserve temporal continuity, as videos are typically partitioned into discrete clips, resulting in visual discontinuities that disrupt the overall video context. To address this challenge, fast-forwarding algorithms have been proposed to efficiently extract meaningful segments while maintaining temporal coherence \cite{ffnet,silva2018weighted,vsum_rout,Ramos2016,Silva2018}.

In parallel, supervised video summarization approaches have leveraged deep neural networks to model temporal dependencies and generate summaries, often relying on annotated human-generated ground truth \cite{stvt,dsnet,a2summ,vasnet,maam,rrstg,msva,hsa,vjmht,dmasum}. Recent advances focus on incorporating attention mechanisms to improve both summarization quality and computational efficiency. For example, models such as VASNet~\cite{vasnet}, SUM-GDA~\cite{sum-gda}, and CA-SUM~\cite{CA-SUM} progressively enhance attention-based summarization by emphasizing efficiency, diversity, and uniqueness, respectively. Transformer-based architectures like DSNet \cite{dsnet} and PGL-SUM~\cite{PGLSUM} refine shot localization and summary precision through attention-driven learning, while VJMHT \cite{vjmht} further boosts performance by capturing semantic similarities across related videos. 

Reinforcement Learning (RL) has demonstrated considerable success across a range of complex tasks, motivating its adoption in the domain of video summarization \cite{Li_2021_WACV,zhou2018deep,ramos2022text,ramos2020straight}. Zhou et al. \cite{zhou2018deep} introduced an end-to-end unsupervised RL framework for video summarization, which optimizes a reward function designed to promote both diversity and representativeness in the generated summaries. Building upon this approach, Li et al. \cite{Li_2021_WACV} incorporated an additional semantic-level reward to further enhance the unsupervised RL model, thereby improving the semantic coherence of the summarized content.

Recent advances in self-supervised learning \cite{barlow,byol,simclr} have demonstrated remarkable effectiveness in video summarization tasks \cite{sspvs,ma,trim}. However, the majority of these approaches predominantly rely on transformer architectures \cite{sspvs} or auto-encoder frameworks with attention mechanisms \cite{ma}. In contrast, recent progress in supervised video summarization \cite{csta} has highlighted the potential of convolutional layers to effectively capture spatio-temporal relationships, achieving state-of-the-art performance on benchmark datasets. Building on this insight, \cite{trim} recently proposed a simple one-dimensional convolutional neural network that achieves competitive results without relying on annotations or complex architectural designs. In this work, we extend that framework by integrating self-supervised learning with reinforcement learning, thereby achieving state-of-the-art performance while maintaining computational efficiency.

\section{Proposed Method}\label{sec:method}

In this section, we propose a two-stage self-supervised reinforcement learning framework for video summarization, motivated by the recent success of TRIM \cite{trim}. Although our approach adopts the same first stage, a self-supervised pretraining stage with a network architecture similar as the one in TRIM method, it diverges significantly in the second stage. Specifically, our reinforcement learning paradigm optimizes a total reward function that captures both the diversity and representativeness of the selected video frames. In this sense, our approach has a partial similarity to the one introduced in \cite{zhou2018deep}. In contrast to TRIM \cite{trim}, which computes the loss based on importance scores from all video frames, our method restricts reward computation to the subset of selected indices $\mathcal{S}$ during training. Furthermore, the proposed framework is computationally efficient and requires minimal hyperparameter tuning; the sole hyperparameter, $\lambda$, serves as a scaling factor to balance the contributions of the individual reward components. Besides, Fig.~\ref{fig:lambda_ablation} (discussed in Section~\ref{ssec:ablation_studies}) presents a study on the influence of $\lambda$ showing that our proposal is stable on $\lambda$, also for different datasets.
\begin{figure*}[htbp]
  \centering
  \includegraphics[width=0.9\textwidth]{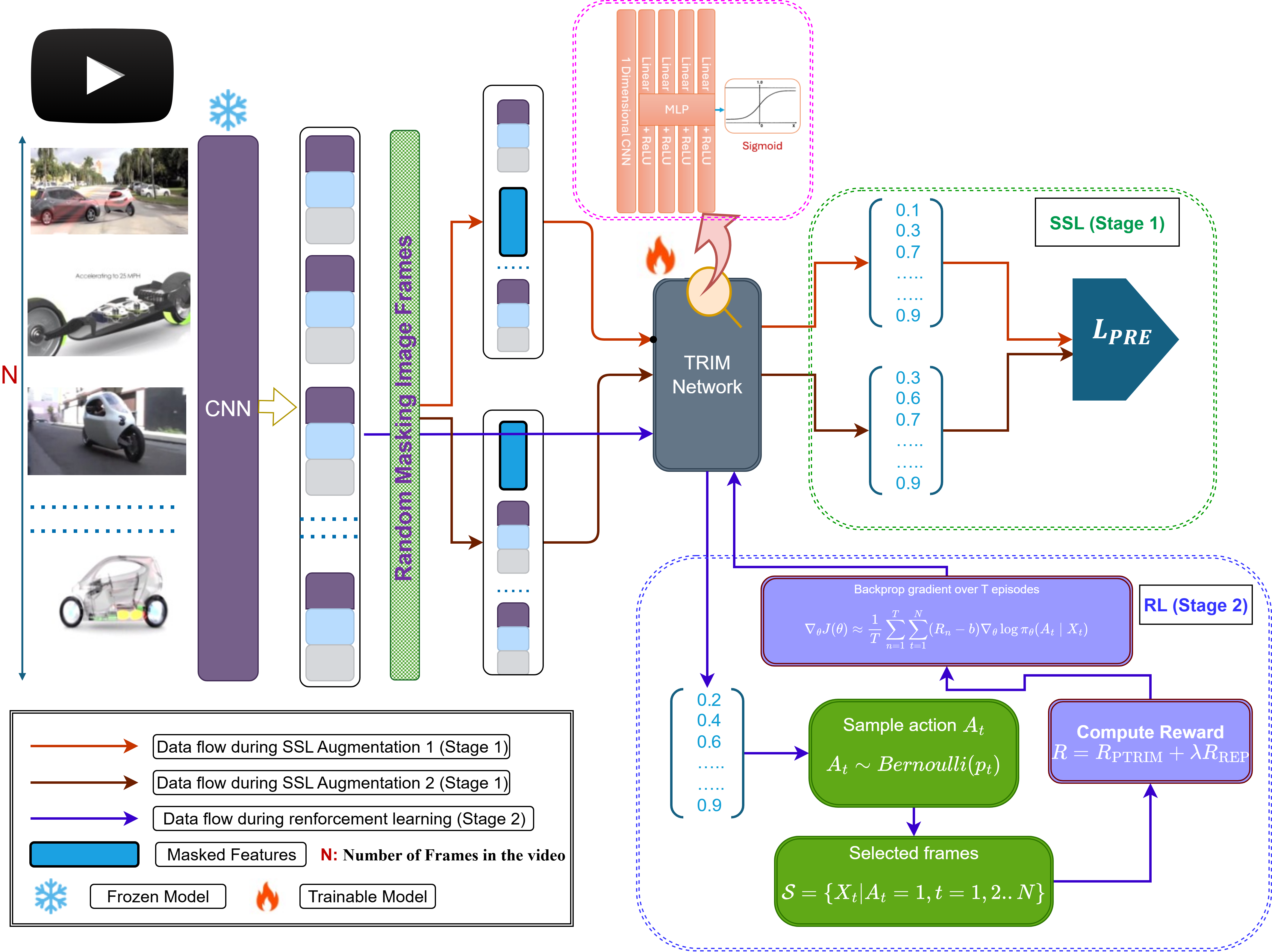}
  \caption{Overview of the proposed two-stage TRIMMER method. The first stage is indicated using orange and brown arrows while the second reinforcement learning stage is indicated using purple arrows.}
  \label{fig:ssl_overview}
\end{figure*}

\subsection{Proposed Pipeline}
We begin by outlining the proposed strategy employed in our method and summarized in Fig.~\ref{fig:ssl_overview}. Our framework employs a two-stage learning pipeline for video summarization. For the sake of clarity, we follow a similar notation as in~\cite{trim}. Given a video $V$ consisting of $N$ frames, namely $\{I_t\}_{t=1}^N$, we extract frame-level features using a frozen convolutional neural network (CNN) $f$, where $X_t = f(I_t)$ denotes the features or embedding of frame $I_t$. 
\begin{itemize}
    \item In the first stage (detailed in Section~\ref{sec:ssl-pre-training}), the network is trained using a self-supervised objective to learn frame-level importance scores $p_t$, for each frame $I_t$. This first stage is indicated using orange and brown arrows in Fig.~\ref{fig:ssl_overview}.
    \item These learned representations are then further refined in the second stage (detailed in Sections~\ref{subc:reinforcement_learning} to \ref{sec:regopt}) using a reinforcement learning approach, which optimizes the summarization quality based on diversity and representativeness. This reinforcement learning stage is indicated using purple arrows and boxes in Fig.~\ref{fig:ssl_overview}.
\end{itemize}
The final summary is constructed by selecting frames based on the importance scores $\{p_t\}_{t=1}^N$ learned by the network (detailed in Section~\ref{sec:summarygen}).

\subsection{Pairwise Temporal Relative Information}
\label{ssec:ptri}
Motivated by the effectiveness of the PTRI metric introduced in TRIM~\cite{trim}, we incorporate it into our proposed diversity reward as a means of quantifying temporal variation in information across video frames. Specifically, PTRI captures the relative change in entropy between consecutive time steps and is defined as follows:
\begin{equation}
    \Delta_t = \left| \frac{H_t - H_{t-1}}{H_t} \right|,
    \label{eq:ptri}
\end{equation}
where $H_t$ denotes the entropy of the feature probability distribution at time step $t$. More precisely, the entropy is computed as:
\begin{equation}
    H_t = H(D_t) = - \sum_{j=1}^{d} D_t^j \log D_t^j,
\end{equation}
where $D_t$ denotes the probability distribution computed from the features $X_t$, $D_t^j$ represents the $j$-th element of the distribution $D_t$, and $d$ denotes the dimensionality of the feature space. This metric provides a principled measure of the frame-wise informational dynamics, which is used as part of our summarization objective.

\subsection{Network Architecture}

Building upon the framework introduced in TRIM~\cite{trim}, we leverage the computationally efficient yet effective network architecture proposed, which has demonstrated superior performance over more computationally intensive models, including RNN-based and Transformer-based approaches~\cite{zhou2018deep,a2summ,sspvs,csta}. This architecture is designed to efficiently model spatio-temporal relationships in video data without relying on recurrent structures, attention mechanisms, or generative adversarial frameworks~\cite{dpplstm,zhou2018deep,gansum,gansum_aae,csta,sspvs,clipit,ma}.

The network consists of a single one-dimensional convolutional layer with a kernel size of 3 and a padding size of 1, followed by a sequence of fully connected layers. Each intermediate layer is followed by a ReLU activation, while the final output layer employs a sigmoid function to produce frame-level importance scores $p_t$ at each time step $t$. The specific architecture is defined as follows:\\
\textbf{$ \mathbf{Conv1D[2048} \to \mathbf{1024]} \to \mathbf{FC[1024} \to \mathbf{512]} \to \mathbf{FC[512} \to \mathbf{256]} \to \mathbf{FC[256} \to \mathbf{1]}$}\\
Notice that each layer progressively reduces the feature dimensionality, enabling the network to capture temporal dependencies while maintaining computational efficiency.

\subsection{Stage 1: Self-Supervised Pre-training}
\label{sec:ssl-pre-training}
Inspired by prior work \cite{trim,ma,sspvs}, we propose a Self-Supervised Learning (SSL) framework as the initial pretraining stage of our approach. Unlike \cite{ma,sspvs}, which employ computationally intensive architectures, our method utilizes a simpler encoder design, motivated by findings from recent contrastive and SSL methods \cite{barlow,simclr,byol} that demonstrate effective representation learning without reliance on auto-encoders or transformer-based models. This suggests that a simple encoder, when combined with a well-designed loss function, can produce robust feature representations.

Specifically, we adopt the loss function introduced in \cite{trim}, which incorporates a Pearson correlation-based term designed to enhance consistency and robustness by maximizing the agreement between two noise-augmented views of the same video representation. Furthermore, we followed \cite{trim} to increase the standard deviation of the importance scores to accentuate the differences across augmented views. This encourages the network to learn noise-invariant representations by requiring it to reconcile larger discrepancies between the augmented inputs, thereby strengthening the robustness of the learned embeddings.

We propose a data augmentation strategy following \cite{trim} that enhances the robustness of our model by randomly masking a percentage $m\%$ of frames within the feature set $X$ extracted from a video $V$. Specifically, the masked frames in the resulting feature set $X'$ (referred to as Augmentation View 1 in Fig.~\ref{fig:ssl_overview}) are replaced with zero vectors. To create a complementary augmented view, this masking process is independently repeated to generate a second feature set $X''$ (Augmentation View 2 in Fig.~\ref{fig:ssl_overview}).

The masking ratio $m$ is sampled between the interval \([15\%, 50\%]\). The influence of different masking ratios on model performance is systematically examined in the ablation studies (Section~\ref{ssec:ablation_studies}).

Both augmented feature sets, $X'$ and $X''$, are then processed through our network, which outputs corresponding frame-level importance score or probability vectors $\mathbf{p_t}' = [p_1', p_2', \dots, p_N']$ and $\mathbf{p_t}'' = [p_1'', p_2'', \dots, p_N'']$. Each element $p_t'$ and $p_t''$ represents the predicted probability of selecting the $t$-th frame, enabling the model to learn noise-invariant and robust representations by maximizing agreement between these two augmented views.

The loss function for the proposed self-supervised pre-training is defined as:
\begin{equation}\label{eq:lossPre}
    L_\text{PRE} = L_\text{CORR} + \nu (L_\text{SD1} + L_\text{SD2}) ,
\end{equation} 
where
\begin{equation}
L_\text{CORR} = 1 - \frac{\sum_{t=1}^{N} (p_t' - \bar{p}')(p_t'' - \bar{p}'')}{\sqrt{\sum_{t=1}^{N} (p_t' - \bar{p}')^2} \cdot \sqrt{\sum_{t=1}^{N} (p_t'' - \bar{p}'')^2}}.
\end{equation}
Let $\bar{p}' = \frac{1}{N} \sum_{t=1}^{N} p_t'$ and $\bar{p}'' = \frac{1}{N} \sum_{t=1}^{N} p_t''$ denote the mean of the vectors $p_t'$ and $p_t''$, respectively. Finally, we define
\begin{equation}
        L_\text{SD1} = \frac{1}{\sqrt{\frac{1}{N}\sum\limits_{t=1}^{N}(p'_t - \bar{p'})^2}} 
, \;
        L_\text{SD2} = \frac{1}{\sqrt{\frac{1}{N}\sum\limits_{t=1}^{N}(p''_t - \bar{p''})^2}}.
\end{equation}
The hyperparameter $\nu\geq 0$ in~\eqref{eq:lossPre} is empirically set to $0.005$ for the \textsc{TVSum} dataset and $0.1$ for the \textsc{SumMe} dataset, based on the results of our ablation studies presented in Section~\ref{ssec:ablation_studies}. For a detailed analysis of the effects of both the masking ratio and $\nu$, we refer the reader to Figs.~\ref{fig:mask_nu_summe} and~\ref{fig:mask_nu_tvsum}.

\paragraph{Training and Optimization.} The proposed network was trained for 90 epochs using the ADAM optimizer \cite{adam_optimizer} with a learning rate of $10^{-5}$ and a weight decay of $10^{-5}$. Training was conducted using the proposed loss function $L_\text{PRE}$ on both the SUMME~\cite{summe_dataset} and TVSUM~\cite{tvsum_dataset} datasets.

\subsection{Stage 2: Reinforcement Learning Paradigm}
\label{subc:reinforcement_learning}
Having obtained robust feature representations through self-supervised pretraining in Stage 1, we now transition to the reinforcement learning phase to fine-tune the network for the task of video summarization. The network, initialized with weights from Stage 1, is trained to predict an importance score for each frame, which is interpreted as the probability of selecting that frame for inclusion in the final summary. These probabilities are then used to sample binary actions from a Bernoulli distribution,  generating a sparse matrix that determines the actions of selecting frames in a video.

In brief, let $A_t$ denote the action of selecting the frame at time $t$, and $\mathcal{S}$ represent the subset of selected frames from a video $V$. To optimize the selection policy, we employ the REINFORCE algorithm~\cite{reinforce_algo} to maximize the expected cumulative (discounted) reward. Furthermore, we apply the regularization techniques introduced in~\cite{zhou2018deep} to encourage diversity and temporal coherence in the generated summaries.

Our reinforcement learning framework for video summarization is grounded in two key principles: \textit{diversity} and \textit{representativeness}, as originally introduced by \cite{zhou2018deep}. To capture diversity, we incorporate the PTRI metric (defined by \eqref{eq:ptri} in Section~\ref{ssec:ptri} above) into the proposed reward function $R_\text{PTRIM}$, defined in \eqref{eq:reward-ptrim} below. This function encourages the selection of frame pairs that exhibit high pairwise information gain —interpreted as surprise or uncertainty— thus encouraging the selection of frames that correspond to semantically distinct content within the video.

To ensure representativeness, we propose a novel reward function $R_\text{REP}$ (defined in \eqref{eq:reward-rep} below), which quantifies how well the selected frames summarize the video content. This is achieved by formulating the problem as a k-medoids clustering task~\cite{tvsum_dataset}, not in the conventional Euclidean space, but in an information-theoretic space based on entropy. For a comprehensive description of our reward design, please refer to Sections~\ref{ssec:reward_ptrim} and~\ref{ssec:reward_rep}.

The cumulative reward for the proposed reinforcement learning framework can be summarized as follows:
\begin{equation}\label{eq:reward}
    R = R_{\mathrm{PTRIM}} + \lambda R_{REP}.
\end{equation}
Here, $\lambda\geq 0$ is a scaling factor that normalizes the rewards to ensure that no single reward disproportionately influences the overall reward maximization. For a comprehensive ablation study on $\lambda$ please refer to Section \ref{ssec:ablation_studies} and Fig.~\ref{fig:lambda_ablation}.

\subsubsection{Pairwise Temporal Relative Information Maximization Reward}
\label{ssec:reward_ptrim}
We can maximize the reward for pairwise temporal relative information for the agent using the PTRI metric as defined earlier but on the set of selected frames $\mathcal{S}$ based on the action $A_t$ taken by the agent at time $t$ for the video $V$. This reward is defined as
\begin{equation}\label{eq:reward-ptrim}
    R_{\mathrm{PTRIM}} = \frac{1}{|\mathcal{S}| - 1}\sum_{t\in\mathcal{S}} \Delta_t
\end{equation}
where $\Delta_t$ is the PTRI metric as defined in equation \eqref{eq:ptri} and $|\mathcal{S}|$ denotes the cardinal of the set of selected indices $\mathcal{S}$.

\subsubsection{Representative Reward}
\label{ssec:reward_rep}
The representative reward is designed to evaluate how effectively the generated summary captures and represents the key content of the original video. To achieve this, we define the representativeness of a video summary in terms of the k-medoids clustering problem, following the approach introduced by Gygli, Grabner, and Van Gool (2015)\cite{tvsum_dataset}. This reward allows us to identify a subset of video segments that best represent the overall content by minimizing the dissimilarity between selected segments (which, by analogy, we will denote as medoids) and the rest of the video. Specifically, we aim for the agent to select a set of medoids that minimizes the difference in entropy between each video frame and its nearest medoid. Therefore, $R_{REP}$ is defined as
\begin{equation}\label{eq:reward-rep}
    R_{\text{REP}} = \exp\left(-\frac{1}{N}\sum_{t=1}^{N} \min_{t' \in \mathcal{S}} \left| H_t - H_{t'} \right| \right).
\end{equation}

\subsection{Training with Policy Gradient}

We adhere to the training and optimization techniques introduced in \cite{zhou2018deep}.
Our goal is to train a policy \(\pi_{\theta}\) that maximizes the expected reward from selected actions over a sequence:
\[
J(\theta) = \mathbb{E}_{p_{\theta}(A_{1:N})}[R]
\]
where \(\theta\) are the policy parameters, \(A_{1:T}\) denotes the sequence of actions, and \(R\) is the reward.

The gradient of this objective is computed using the REINFORCE algorithm, which is well-suited for scenarios where the reward signal is only available at the sequence (episodic) level:
\begin{equation}
\nabla_{\theta} J(\theta) = \mathbb{E}_{p_{\theta}(A_{1:N})} \left[ R \sum_{t=1}^N \nabla_{\theta} \log \pi_{\theta}(A_t \mid X_t) \right].
\end{equation}
Here, \( X_t \) denotes the input feature vector at time step \( t \), and the policy \( \pi_{\theta} \) maps these features to the probability of selecting action \( A_t \).

Since the expectation is intractable, the gradient is approximated using Monte Carlo sampling over \( T \) episodes:
\begin{equation}
\nabla_{\theta} J(\theta) \approx \frac{1}{T} \sum_{n=1}^T \sum_{t=1}^N R_n \nabla_{\theta} \log \pi_{\theta}(A_t \mid X_t)
\end{equation}
where \( R_n \) denotes the episodic reward obtained in the \( n \)-th sample trajectory.

To reduce the variance of the estimator and improve training stability, a scalar baseline \( b \) is subtracted from the reward:
\begin{equation}
\nabla_{\theta} J(\theta) \approx \frac{1}{T} \sum_{n=1}^T \sum_{t=1}^N (R_n - b) \nabla_{\theta} \log \pi_{\theta}(A_t \mid X_t)
\end{equation}
The baseline is implemented as a moving average  and T is set to 5 following \cite{zhou2018deep}.

\subsection{Regularization and Optimization}\label{sec:regopt}

To encourage desirable policy behavior, such as maintaining a target selection ratio \(\epsilon\), we include a regularization term:
\begin{equation}
\label{eq:regularization}
L_{\text{percentage}} = \left\| \frac{1}{N} \sum_{t=1}^N p_t - \epsilon \right\|^2
\end{equation}
where \(p_t\) denotes the importance score at time \(t\) and $\epsilon$ is set to $0.5$ following \cite{zhou2018deep}.

The complete parameter update rule integrates the expected reward objective with regularization terms:
\[
\theta \leftarrow \theta - \alpha \nabla_{\theta} \left( -J + \beta L_{\text{percentage}} + L_{\text{weight}} \right).
\]
Here, \(\alpha\) is the learning rate set to $10^{-5}$, \(L_{\text{weight}}\) is an additional regularization term such as L2 weight decay set to $10^{-5}$, and $\beta$ controls the contribution of $L_{\text{percentage}}$ set to 0.01 following \cite{zhou2018deep}.

We optimize the parameters of the proposed network using the Adam \cite{adam_optimizer} optimizer for 60 epochs adhering to the optimization protocol introduced in \cite{zhou2018deep}.

\subsection{Summary Generation}\label{sec:summarygen}
We adopt a video summarization procedure consistent with prior works~\cite{csta,trim,zhou2018deep,a2summ}. During inference, the proposed network predicts an importance score $p_t$ for each frame $I_t$ in the video $V$. To generate the final summary of $V$, we first apply Kernel Temporal Segmentation (KTS)~\cite{kts} to divide the video into temporally coherent segments. For each segment, we compute the mean importance score and then employ dynamic programming to solve a 0/1 Knapsack problem~\cite{tvsum_dataset}, selecting a subset of segments such that the total duration of the summary does not exceed 15\% of the original video length.

\section{Experimental Results}\label{sec:experiments}
This section presents the experiments supporting the main findings of this work, providing a comprehensive evaluation of the proposed approach across multiple datasets and metrics to demonstrate its efficacy, efficiency, and practical applicability.

\begin{table*}[htbp]
    \centering
    \renewcommand{\arraystretch}{1.2}
    \scriptsize
    \caption{Comparative evaluation of the proposed  framework TRIMMER against existing state-of-the-art methods using ranking correlation metrics—Spearman’s \(\rho\) and Kendall’s \(\tau\). Methods categorized as unsupervised or self-supervised are shaded in gray. The columns \textbf{SUP} and \textbf{SSL} denote whether a method employs supervised or self-supervised learning, respectively; \textbf{FT} indicates if the network has been fine-tuned after self-supervision; \textbf{ATTN} specifies the use of attention mechanisms in the network; and \textbf{GFLOPs} reports the computational complexity measured in gigaflops.}
    \label{tab:sota_comparisonRL}
    \begin{tabular}{lcc|cc|c|c|c|c|c}
        \toprule
        \multirow{2}{*}{Method} & \multicolumn{2}{c|}{TVSum} & \multicolumn{2}{c}{SumMe} & SUP & SSL & FT & ATTN & GFLOPs \\
        & $\tau$ & $\rho$ & $\tau$ & $\rho$ \\
        \midrule
        Human & 0.177 & 0.204 & 0.212 & 0.212 & - & - & - & - & -\\
        Random & 0.000 & 0.000 & 0.000 & 0.000 & - & - & - & - & - \\
        \midrule
        dppLSTM \cite{dpplstm} & 0.042 & 0.055 & -0.026 & -0.031 & \textbf{\checkmark} & \xmark & \xmark & \xmark & - \\
        \rowcolor{gray!20}
        DR-DSN \cite{zhou2018deep} & 0.020 & 0.026 & 0.043 & 0.050 & \xmark & \xmark & \xmark & \xmark & - \\
        DR-DSN$_{2000}$\cite{zhou2018deep} & 0.152 & 0.198 & -0.016 & -0.022 & - & - & - & \xmark & - \\
        \rowcolor{gray!20}
        GANSUM \cite{gansum} & 0.024 & 0.032 & -0.010 & -0.012 & \xmark & \xmark & \xmark & \xmark & - \\
        \rowcolor{gray!20}
        CSNet\cite{csnet} & 0.025 & 0.034 & - & - & \xmark & \xmark & \xmark & \textbf{\checkmark} & - \\
        \rowcolor{gray!20}
        GANSUM+AAE\cite{gansum_aae} & -0.047 & -0.062 & -0.018 & -0.023 & \xmark & \xmark & \xmark & \textbf{\checkmark} & -\\
        DAC\cite{dac} & 0.058 & 0.065 & 0.063 & 0.059 & \textbf{\checkmark} & \xmark & \xmark & \textbf{\checkmark} & - \\
        \rowcolor{gray!20}
        GANSUM+GL+RPE \cite{gansum_gl_rpe} & 0.064 & 0.084 & - & -  & \xmark & \xmark & \xmark & \xmark & - \\
        \rowcolor{gray!20}
        CSNet+GL+RPE\cite{gansum_gl_rpe} & 0.070 & 0.091 & - & -  & \xmark & \xmark & \xmark & \xmark & -\\
        HSA-RNN\cite{hsa} & 0.082 & 0.088 & 0.064 & 0.066 & \textbf{\checkmark} & \xmark & \xmark & \xmark & -\\
        DAN\cite{dan} & 0.071 & 0.099 & - & - & \textbf{\checkmark} & \xmark & \xmark & \textbf{\checkmark} & -\\
        HMT\cite{hmt} & 0.096 & 0.107 & 0.079 & 0.080 & \textbf{\checkmark} & \xmark & \xmark & \textbf{\checkmark} & - \\
        VJMHT\cite{vjmht} & 0.097 & 0.105 & 0.106 & 0.108 & \textbf{\checkmark} & \xmark & \xmark & \textbf{\checkmark} & 56.5G \\
        STVT\cite{stvt} & 0.1 & 0.131 & - & - & \textbf{\checkmark} & \xmark & \xmark & \textbf{\checkmark} & -\\
        DSNet-AB\cite{dsnet} & 0.108 & 0.129 & 0.051 & 0.059 & \textbf{\checkmark} & \xmark & \xmark & \textbf{\checkmark} & 4.14G \\
        DSNet-AF\cite{dsnet} & 0.113 & 0.138 & - & - & \textbf{\checkmark} & \xmark & \xmark & \textbf{\checkmark} & 3.8G \\
        \rowcolor{gray!20}
        MA \cite{ma} & 0.110 & 0.144 & 0.046 & 0.057 & \xmark  & \textbf{\checkmark} & \textbf{\checkmark} & \textbf{\checkmark} & - \\
        CLIP-It\cite{clipit} & 0.108 & 0.147 & - & - & \textbf{\checkmark} & \xmark & \xmark & \textbf{\checkmark} & -\\
        iPTNet\cite{ipnet} & 0.134 & 0.163 & 0.101 & 0.119 & \textbf{\checkmark} & \xmark & \xmark & \textbf{\checkmark} & -\\
        A2Summ \cite{a2summ} & 0.137 & 0.165 & 0.108 & 0.129 & \textbf{\checkmark} & \xmark & \xmark & \textbf{\checkmark} & -\\
        VASNet \cite{vasnet} & 0.160 & 0.170 & 0.160 & 0.170 & \textbf{\checkmark} & \xmark & \xmark & \textbf{\checkmark} & 4.6G\\
        MAAM\cite{maam} & 0.179 & 0.236 & - & - & \textbf{\checkmark} & \xmark & \xmark & \textbf{\checkmark} & -\\
        VSS-Net \cite{vssnet} & 0.190 & 0.249 & - & - & \textbf{\checkmark} & \xmark & \xmark & \textbf{\checkmark} & -\\
        DMASum \cite{dmasum} & 0.203 & 0.267 & 0.063 & 0.089 & \textbf{\checkmark} & \xmark & \xmark & \textbf{\checkmark} & -\\
        RR-STG \cite{rrstg} & 0.162 & 0.212 & 0.211 & 0.234 & \textbf{\checkmark} & \xmark & \xmark & \textbf{\checkmark}  & - \\
        MSVA \cite{msva} & 0.190 & 0.210 & 0.200 & 0.230 & \textbf{\checkmark} & \xmark & \xmark & \textbf{\checkmark} & 11.62G \\
        SSPVS \cite{sspvs} & 0.181 & 0.238 & 0.192 & 0.257 & \xmark & \textbf{\checkmark}  & \textbf{\checkmark} & \textbf{\checkmark} & 88.44G \\
        CSTA \cite{csta} & 0.194 & 0.255 & 0.246 & 0.274 & \textbf{\checkmark} & \xmark & \xmark & \textbf{\checkmark} & 31.46G \\
        \rowcolor{gray!20}
        TRIM \cite{trim} & 0.181 & 0.228 & 0.127 & 0.140 & \xmark & \textbf{\checkmark} & \textbf{\checkmark} & \xmark & \textbf{3.27G}\\
        \midrule
        \rowcolor{gray!20}
        \textbf{TRIMMER} & \textbf{0.195} & \textbf{0.255} & \textbf{0.135} & \textbf{0.151} & \xmark & \textbf{\checkmark} & \textbf{\checkmark} & \xmark & \textbf{3.27G}\\
        \bottomrule
    \end{tabular}
\end{table*}

\begin{figure}
\vskip 0.2in
\begin{center}
\centerline{\includegraphics[width=0.75\textwidth]{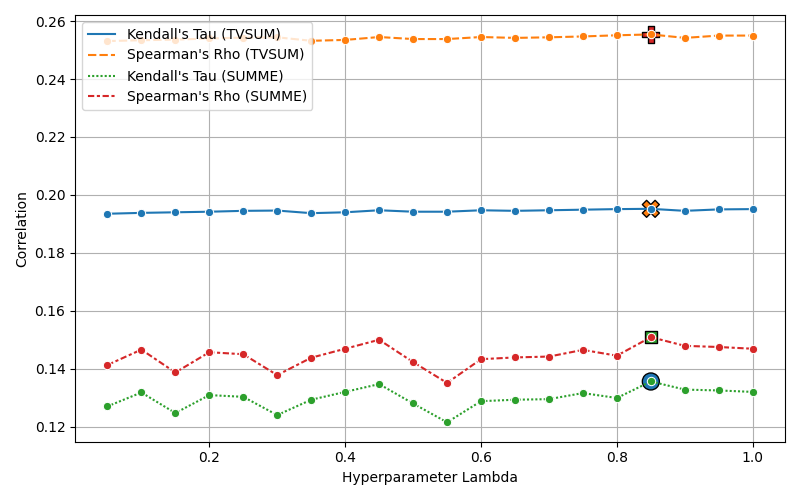}}
\caption{Ablation study on the influence of hyperparameter $\lambda$ trained with our proposed reward ($R = R_{REP} + \lambda R_{\mathrm{PTRIM}}$) function after self-supervision pre-training on TVSUM\cite{tvsum_dataset} and SUMME\cite{summe_dataset} datasets.}
\label{fig:lambda_ablation}
\end{center}
\vskip -0.2in
\end{figure}
\raggedbottom

\begin{table}[h]
    \centering
    \small
    \caption{Ablation study on the different terms of the reward  $R = R_{REP} + \lambda R_{\mathrm{PTRIM}}$ on SumMe\cite{summe_dataset} and TVSum\cite{tvsum_dataset} datasets (with $\lambda$ fixed to  $0.85$ according to the stability on $\lambda$ of our proposal  and discussion summarized in Fig.~\ref{fig:lambda_ablation}).}
    \begin{tabular}{lcc|cc}
        \toprule
        \textbf{Loss} & \multicolumn{2}{c}{\textbf{TVSum}} & \multicolumn{2}{c}{\textbf{SumMe}} \\
        & $\tau$ & $\rho$ & $\tau$ & $\rho$ \\
        \midrule
        $R_{\mathrm{PTRIM}}$ & 0.186 & 0.244 & 0.126 & 0.140 \\
        $R_\text{REP}$ & 0.190 & 0.250 & 0.132 & 0.148 \\
        $R_{\mathrm{PTRIM}} + \lambda R_\text{REP}$ & 0.195 & 0.255 & 0.135 & 0.151 \\
        \bottomrule
    \end{tabular}
    \label{tab:ablation}
\end{table}

\subsection{Evaluation Methods}
\label{secc:eval_methods}
Several prior studies have emphasized that model performance can vary significantly due to randomness in the training process, particularly given the stochastic nature of modern optimization algorithms. To ensure a robust and reproducible evaluation, we adopt a rigorous experimental protocol consistent with recent works~\cite{csta,maam,trim}. Specifically, each experiment is repeated ten times, with each run involving five-fold cross-validation using non-overlapping test sets across folds. The final reported performance reflects the average across these ten independent runs.

We evaluate the effectiveness of the proposed method using rank-based correlation coefficients---namely, Kendall’s $\tau$ and Spearman’s $\rho$---to measure the agreement between the predicted importance scores and human-annotated ground truth. This choice is motivated by recent state-of-the-art approaches in video summarization~\cite{csta,maam,vssnet,sspvs,ipnet,clipit,trim}, which have moved away from the traditional F1-score-based evaluation.
Indeed, historically, F1 score has been the dominant metric for assessing performance on standard benchmarks such as \textsc{SumMe} and \textsc{TVSum} datasets, typically based on a single test run. However, recent research has challenged this practice~\cite{rethinking}, demonstrating that F1 scores can be artificially inflated when models preferentially select a large number of short segments over fewer, longer ones---an artifact of the fixed-length constraint imposed on the summary. This limitation has been documented in several recent works~\cite{rethinking,maam,csta,trim}. Furthermore, methods that report high F1 scores, such as~\cite{zhou2018deep}, have been shown to perform poorly when evaluated using rank-based metrics~\cite{trim}. Notably, it has also been observed that even randomly generated predictions can yield deceptively high F1 scores, while their rank correlation with the ground truth remains close to zero~\cite{rethinking,trim}.

\subsection{Comparison with the state of the art}
We demonstrate the performance of our proposed self-supervised method fine-tuned with reinforcement learning paradigm in Table \ref{tab:sota_comparisonRL}. The results show that the proposed method achieves state-of-the-art results compared to all other unsupervised and self-supervised methods and comparable performance against supervised methods. Let us note that, in all the results presented in Tables \ref{tab:sota_comparisonRL} and \ref{tab:ablation}, $\lambda$ fixed to $0.85$ according to the stability on $\lambda$ of our proposal, the discussion of which is summarized in Fig.~\ref{fig:lambda_ablation} (details in Section~\ref{ssec:ablation_studies}).

\subsection{Ablation studies}
\label{ssec:ablation_studies}
To assess the effectiveness of the individual components of our reward function and the influence of hyperparameter choices, we conduct a series of ablation studies. These experiments are designed to isolate and evaluate the contribution of each reward component and the impact of different hyperparameter settings. Initially, we demonstrate that each constituent of the proposed reward function plays a critical role in enhancing the overall performance of our method. In particular, we observe that the diversity and representativeness terms are complementary, jointly contributing to the state-of-the-art results achieved on SumMe and TVSum datasets. The results presented in Table~\ref{tab:ablation} compare the performance of the individual reward terms $R_\text{PTRIM}$ and $R_\text{REP}$, as well as their combination, $R=R_{\mathrm{PTRIM}}+\lambda R_{REP}$.  

\begin{figure}
\vskip 0.2in
\begin{center}
\centerline{\includegraphics[width=0.75\textwidth]{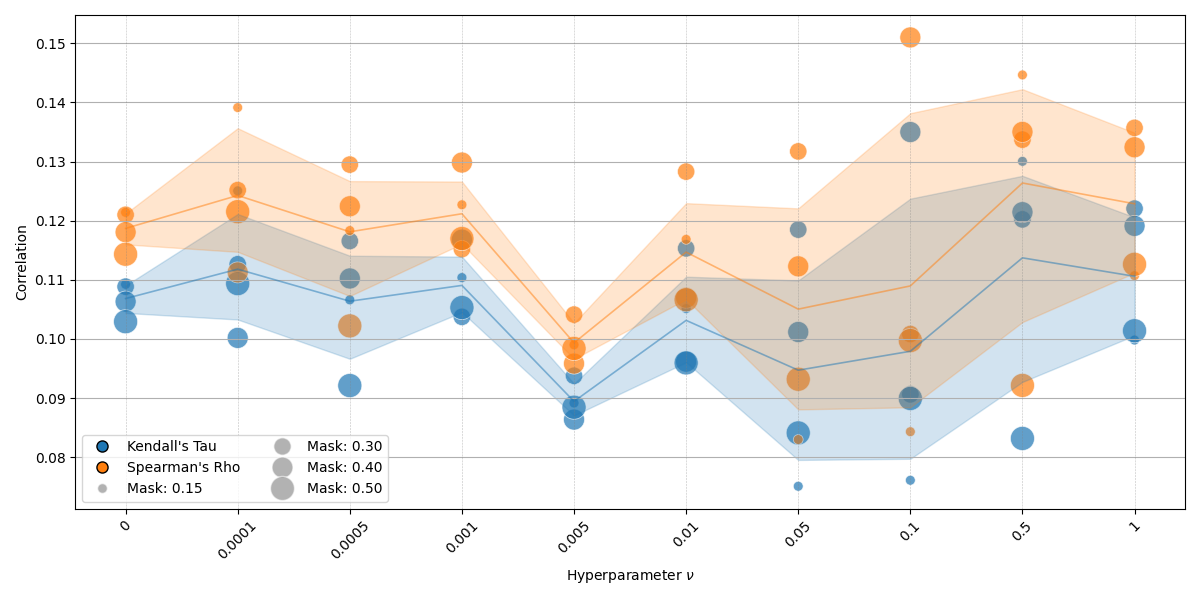}}
\caption{Ablation study of mask ratio m and hyperparameter $\nu$ when trained with our proposed reward function ($R = R_{REP} + \lambda R_{\mathrm{PTRIM}}$) after self-supervised pretraining on SUMME \cite{summe_dataset} dataset. Each correlation type is visually distinguished by a unique combination of color (hue), line style, and marker.}
\label{fig:mask_nu_summe}
\end{center}
\vskip -0.2in
\end{figure}

\begin{figure}
\vskip 0.2in
\begin{center}
\centerline{\includegraphics[width=0.75\textwidth]{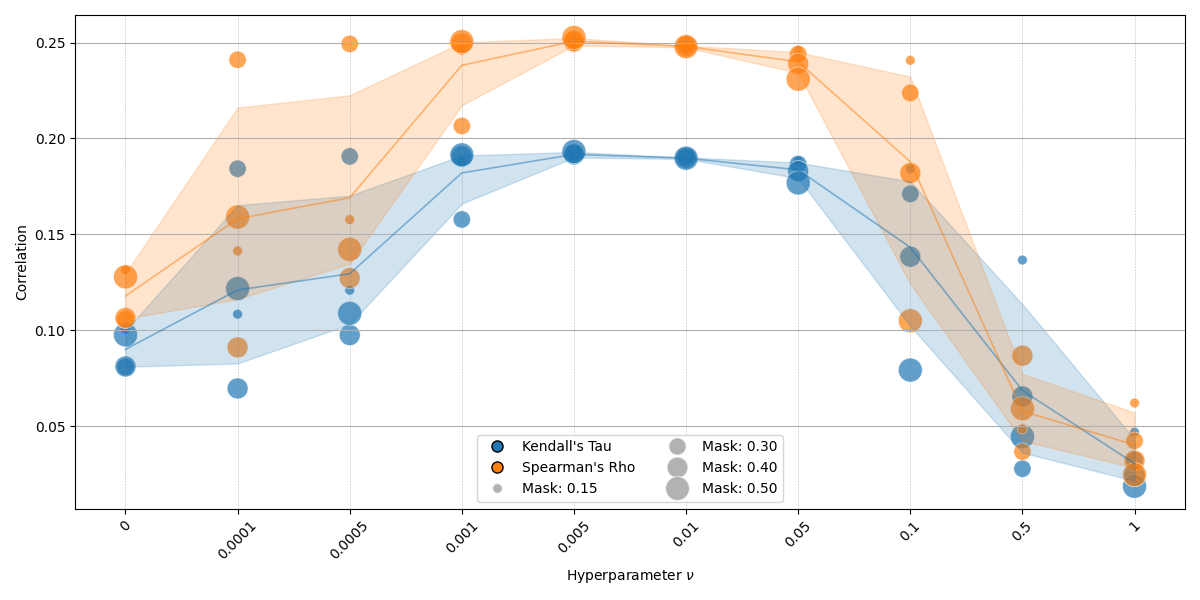}}
\caption{Ablation study of mask ratio m and hyperparameter $\nu$ when trained with our proposed reward function ($R = R_{REP} + \lambda R_{\mathrm{PTRIM}}$) after self-supervised pretraining on TVSUM \cite{tvsum_dataset} dataset. Each correlation type is visually distinguished by a unique combination of color (hue), line style, and marker.}
\label{fig:mask_nu_tvsum}
\end{center}
\vskip -0.2in
\end{figure}

Furthermore, we investigate the impact of the hyperparameter $\nu$ and the masking ratio $m$ used during Stage 1 of self-supervised learning. Specifically, we perform a comprehensive grid search over masking ratios $\{0.15, 0.3, 0.4, 0.5\}$ and $\nu$ values $\{0, 0.0001, 0.0005, 0.001, 0.005, 0.01, 0.05, 0.1, 0.5, 1.0\}$, resulting in a total of 40 experiments for each dataset (\textsc{SumMe} and \textsc{TVSum}). The outcomes of these experiments are illustrated in Fig.~\ref{fig:mask_nu_summe} and Fig.~\ref{fig:mask_nu_tvsum}, corresponding to the \textsc{SumMe} and \textsc{TVSum} datasets, respectively.

\raggedbottom

\raggedbottom

This grid-based ablation analysis enables us to identify optimal hyperparameter configurations for each dataset. Notably, we observe that the best performance on \textsc{TVSum} is achieved with a masking ratio $m = 0.3$ and $\nu = 0.005$, consistent with the hyperparameter settings reported in~\cite{trim}. In contrast, for the \textsc{SumMe} dataset, the optimal configuration diverges from~\cite{trim}, with a masking ratio $m = 0.4$ and $\nu = 0.1$ yielding superior results. These findings underscore the importance of dataset-specific hyperparameter tuning, as the two benchmark datasets differ significantly in key aspects such as video length, content diversity, and dataset composition. Specifically, \textsc{TVSum} contains videos grouped into predefined categories, while \textsc{SumMe} comprises a more diverse set of unstructured videos (further explained in section \ref{ssec:dataset_description}). Additionally, the number of videos and their durations vary considerably between the datasets, necessitating tailored hyperparameter choices for optimal performance.

Finally, we conduct an ablation study on the hyperparameter $\lambda$ in \eqref{eq:reward} during Stage 2 of reinforcement learning, using the optimal hyperparameters identified during Stage 1 self-supervised learning for both datasets. The hyperparameter $\lambda$ plays the role of balancing the contributions of the reward components, particularly because the reward functions operate on different scales—specifically, $R_\text{REP}$ tends to yield higher reward values compared to $R_\text{PTRIM}$. The study, which is summarized in Fig.~\ref{fig:lambda_ablation}, reveals that the optimal value of $\lambda$ is $0.85$ for both datasets, \textsc{SumMe} and \textsc{TVSum}. Let us also note that the performance of our proposal shows stability varying $\lambda$, especially for \textsc{TVSum} dataset.

\section{Additional Analysis and Discussion}\label{sec:additional_analysis}

\subsection{Conceptual Foundations of the TRIMMER Framework}
\label{ssec:intuition}
The intuition behind our proposed self-supervised reinforcement learning framework is grounded in the observation that frame selection for video summarization is inherently a combinatorial optimization problem. Given a video consisting of $N$ frames, the number of possible frame subsets grows exponentially with $N$, making the problem particularly challenging for reinforcement learning methods. Indeed,
\begin{itemize}
\item If a fixed number $k$ of frames is to be selected, the number of possible summaries for a video $V$ is given by the binomial coefficient: $\binom{N}{k} = \frac{N!}{k!(N - k)!}$.

\item If $k$ is not fixed, the total search space comprises all possible subsets of the $N$ frames, i.e., $2^N$ combinations.

\item However, since $N$ varies across videos, a fixed $k$ is not suitable as a global hyperparameter.

\item Selecting a random $k$ per video introduces high variance and increases the risk of overfitting, especially given the combinatorial nature of the problem. This approach would require a prohibitively large number of episodes to converge to an optimal policy.


\item Moreover, training the policy network for only a limited number of epochs with randomly sampled $k$ values tends to produce trivial solutions. For instance, DR-DSN \cite{zhou2018deep} trained their policy network for 60 epochs with randomly assigned $k$ values, resulting in suboptimal summarization performance, as evidenced in Table~\ref{tab:sota_comparisonRL}.
\end{itemize}
To address these limitations, we have proposed a hybrid framework that integrates self-supervised learning with reinforcement learning. In the self-supervised pretraining phase, the network learns a representation for each video based on its visual content, independent of ground-truth labels. This pretraining phase allows the model to assign an initial importance score to each frame, from which an approximate value of $k$ can be inferred adaptively for each video.

By leveraging these learned frame-level scores to guide reinforcement learning, our approach offers several advantages:
\begin{itemize}
    \item[(i)] It reduces the risk of overfitting by constraining the policy space.
    \item[(ii)] It avoids trivial or degenerate solutions often caused by random $k$ initialization.
    \item[(iii)] It enables reproducible training with consistent initialization across seeds and folds.
    \item[(iv)] It mitigates the computational burden typically associated with exploring a large combinatorial action space.
\end{itemize}

\subsection{Computational Efficiency and Complexity Analysis}
\label{ssec:bigo_analysis}
In this section, we outline the computational resources utilized for training our proposed method and present a comprehensive comparison of its computational complexity with closely related approaches. All experiments were conducted using a single NVIDIA TITAN RTX GPU. Our proposed network demonstrates high computational efficiency, achieving a complexity of \textbf{3.27 GFLOPs}, which is significantly lower than that of several state-of-the-art methods, as shown in Table~\ref{tab:sota_comparisonRL}.

Moreover, we provide a detailed analysis of the computational complexity, expressed in Big-O notation, for each individual reward component in our framework. This comparison includes relevant reward/loss components from TRIM \cite{mishra2026trim}, TRIMMER and DRDSN~\cite{zhou2018deep}, enabling a clear understanding of the relative computational costs. Table~\ref{tab:complexity_reward} summarizes the complexity of each reward term in our method alongside those of TRIM \cite{mishra2026trim} and DRDSN, highlighting the efficiency of our design.

\begin{table}[h]
\centering
\footnotesize
\caption{Comparison of computational complexity (in terms of Big-O notation) for each reward function component in TRIMMER with TRIM, and DRDSN~\cite{zhou2018deep}.}
\label{tab:complexity_reward}
\begin{tabular}{lcc}
\toprule
\textbf{Reward/Loss Function} & \textbf{Computational Complexity (Big O)} \\
\midrule
TRIM (Ours)\\
\midrule
$L_\text{PTRIM} = \frac{N-1}{\sum\limits_{t=2}^{N} \, p_t \,\Delta_t}$ & $\mathcal{O}(N)$ \\
$L_\text{PCTRIM} = \frac{N-1}{\sum\limits_{t=2}^{N}\, p_t \, \Gamma_t}$ & $\mathcal{O}(N)$  \\
$W_2^2(a, b) = \min_{\gamma \in \Pi(a, b)} \sum\limits_{i=1}^{N} \sum\limits_{j=1}^{N} \gamma_{ij} \| X_i - p_j X_j \|^2$ & $\mathcal{O}(N^{2})$  \\
$L_\text{SD} = \frac{1}{\sqrt{\frac{1}{N}\sum\limits_{t=1}^{N}(p_t - \bar{p})^2}}$ & $\mathcal{O}(N)$  \\
\midrule
DRDSN \cite{zhou2018deep} \\
\midrule
$R_{\text{div}} = \frac{1}{k(k-1)} \sum_{\substack{i,j \in \mathcal{S} \\ i \ne j}} d(X_i, X_j)$ & $\mathcal{O}(k^{2})$, where $k = |\mathcal{S}| < N$ \\
$R_{\text{rep}} = - \frac{1}{N} \sum_{t=1}^{N} \min_{s \in \mathcal{S}} d(X_t, X_s)$ & $\mathcal{O}(Nk)$, where $k = |\mathcal{S}| < N$ \\
\midrule
TRIMMER (Ours)\\
\midrule
$R_{\mathrm{PTRIM}} = \frac{1}{|\mathcal{S}| - 1}\sum_{t\in\mathcal{S}} \Delta_t$ & $\mathcal{O}(k)$, where $k = |\mathcal{S}| < N$  \\
$R_{\mathrm{REP}} = \exp\left(-\frac{1}{N}\sum_{t=1}^{N} \min_{t' \in \mathcal{S}} \left| H_t - H_{t'} \right| \right)$ & $\mathcal{O}(Nk)$, where $k = |\mathcal{S}| < N$ \\
\bottomrule
\end{tabular}
\end{table}

\section{Conclusion}
This work presents a generalized two-stage self-supervised learning paradigm for video summarization, originally introduced in our earlier framework TRIM \cite{trim} and further extended in this work through the introduction of TRIMMER. The proposed approach improves optimization efficiency by incorporating a reinforcement learning paradigm that leverages entropy-based reward mechanisms to guide the selection of informative and diverse video segments without relying on dataset-specific manual annotations. By combining scalable representation learning with efficient decision-making, the framework enables robust summarization while remaining computationally practical for large-scale video data. Owing to its generalized design and independence from extensive labeled data, the proposed paradigm may also be applicable to a broader range of video understanding tasks where annotated datasets are limited but large volumes of unlabeled video are available, such as  editorial assistance in media broadcasting, video anomaly detection, long-term surveillance analysis, and large-scale CCTV monitoring systems.

\clearpage

\bibliography{references}

\clearpage

\appendix
\section{Appendix}
\subsection{Dataset Description}\label{ssec:dataset_description}

\paragraph{SumMe Dataset} The SumMe dataset \cite{summe_dataset} is a widely used benchmark to evaluate video summarization algorithms. It contains 25 user-generated videos that depict various real-world scenarios such as holidays, sports, and public events. The videos are collected from YouTube and vary in camera perspectives, including static, egocentric, and moving shots. Their duration ranges from 1 to 6 minutes, and they exhibit significant diversity in both visual style and content.

Each video is annotated with multiple human-generated summaries obtained from at least 15 participants who independently select keyframes or keyshots they consider important. These annotations serve as ground truth for evaluation, where models are assessed based on their ability to predict the average frame-level importance derived from human selections. The dataset reflects the inherently subjective nature of video summarization and captures the variability in human judgments regarding salient video content.

\paragraph{TVSum Dataset}
The TVSum dataset \cite{tvsum_dataset}, also known as TV Episode Summarization, is another commonly used benchmark for video summarization. It consists of 50 videos ranging from 2 to 10 minutes in length and is derived from the TRECVid Multimedia Event Detection (MED) dataset \cite{trec_tvsum}. The videos cover 10 event categories—such as changing a vehicle tire, grooming an animal, and making a sandwich—with five videos per category.

Ground truth annotations are provided by 20 crowd-sourced annotators for each video, who assign shot-level importance scores on a scale from 1 to 5. These annotations enable the evaluation of summarization models based on their ability to estimate the average importance of video segments. Compared to SumMe, which focuses on diverse user-generated content and subjective preferences, TVSum provides a more structured and event-centric setting, making it particularly suitable for studying task-oriented video summarization.

\subsection{Empirical Analysis of Individual Reward Terms in TRIMMER}
In this section, we conduct a deeper analysis of the behavior of the individual reward components in our framework and their contribution toward achieving state-of-the-art performance. Specifically, we compare the predicted importance scores obtained from each reward term with the ground truth annotations—defined as the mean importance scores across all annotators—to evaluate whether the learned predictions align with the underlying design intuition of our reward functions. Furthermore, we analyze the temporal progression of reward accumulation over training epochs to illustrate how the two complementary rewards contribute jointly to the multi-objective optimization.

\begin{figure}
\vskip 0.2in
\begin{center}
    \begin{subfigure}[t]{0.49\textwidth}
        \centering
        \includegraphics[width=\linewidth]{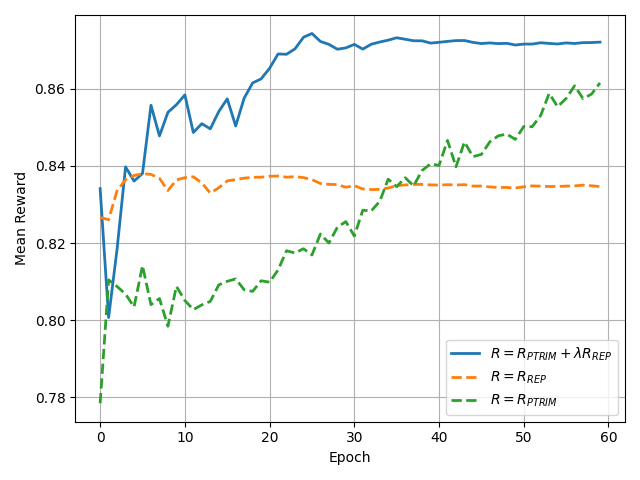}
        \caption{Reward progression on TVSUM~\cite{tvsum_dataset} dataset.}
        \label{subfig:reward_tvsum}
    \end{subfigure}
    \begin{subfigure}[t]{0.49\textwidth}
        \centering
        \includegraphics[width=\linewidth]{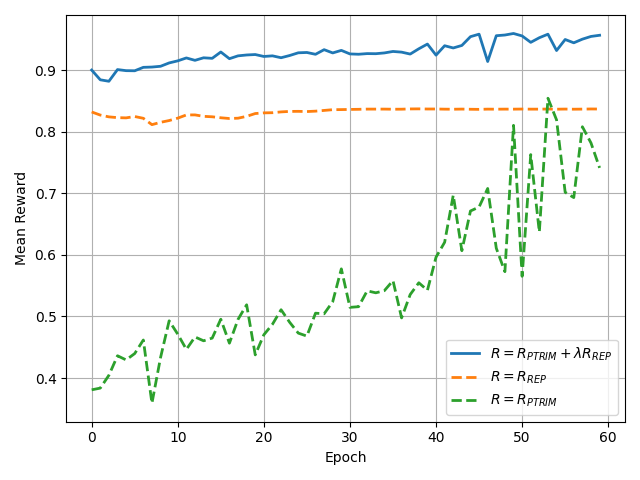}
        \caption{Reward progression on SUMME~\cite{summe_dataset} dataset.}
        \label{subfig:reward_summe}
    \end{subfigure}

\caption{Ablation study showing the progression of mean rewards across all videos during training, evaluated via five-fold cross-validation using a fixed random seed. The comparison includes individual reward components $R_{\mathrm{PTRIM}}$, $R_\text{REP}$, and the cumulative reward defined as $R = R_{\mathrm{PTRIM}} + \lambda R_\text{REP}$, used during the reinforcement learning phase (Stage 2) following pretraining on the TVSUM dataset \cite{tvsum_dataset} (Figure~\ref{subfig:reward_tvsum}) and SUMME dataset \cite{summe_dataset} (Figure~\ref{subfig:reward_summe}). For improved visual clarity, reward values were scaled to facilitate alignment across curves.}

\label{fig:reward_ablation}
\end{center}
\vskip -0.2in
\end{figure}
\raggedbottom

\begin{figure}[h]
    \centering

    \begin{subfigure}[t]{0.47\textwidth}
        \centering
        \includegraphics[width=\linewidth]{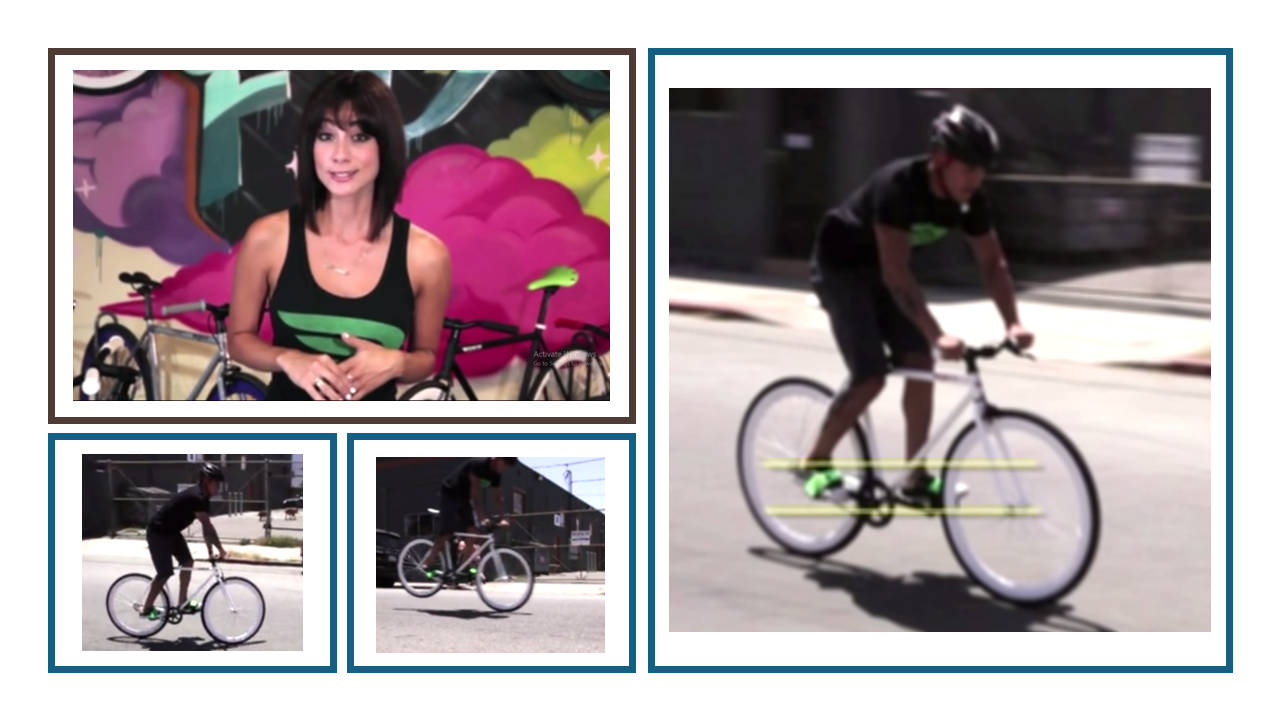}
        \caption{Key frames- Video "eQu1rNs0an0"}
    \end{subfigure}
    \begin{subfigure}[t]{0.5\textwidth}
        \centering
        \includegraphics[width=\linewidth]{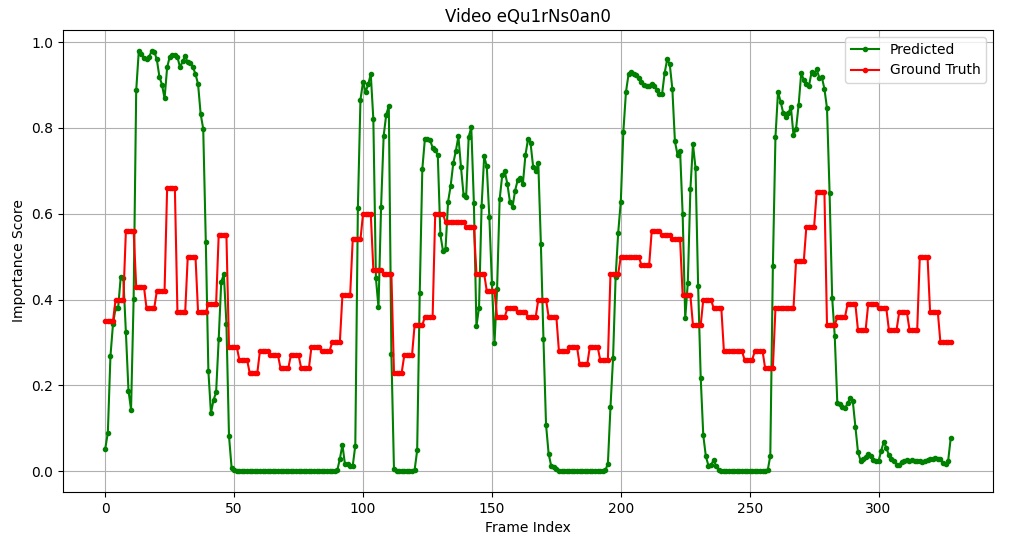}
        \caption{$R_\text{REP}$}
    \end{subfigure}

    \vspace{0.1cm}

    \begin{subfigure}[t]{0.49\textwidth}
        \centering
        \includegraphics[width=\linewidth]{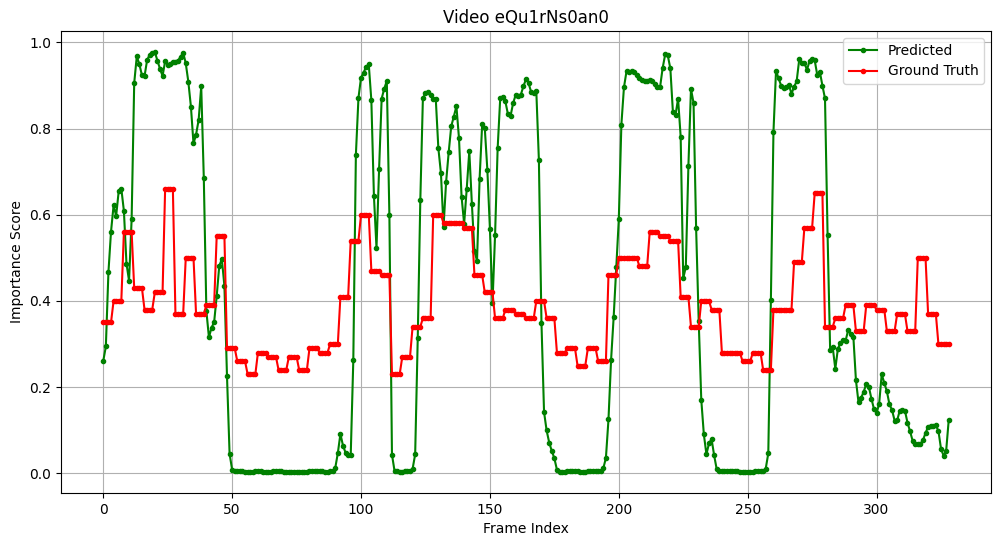}
        \caption{$R_{\mathrm{PTRIM}}$}
    \end{subfigure}
    \begin{subfigure}[t]{0.49\textwidth}
        \centering
        \includegraphics[width=\linewidth]{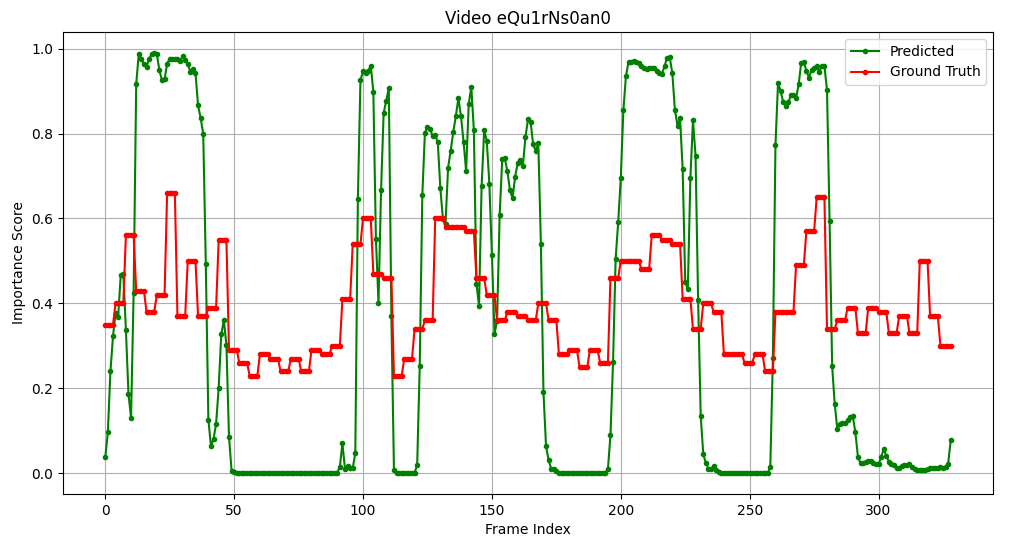}
        \caption{$R_{\mathrm{PTRIM}} + \lambda R_\text{REP}$}
    \end{subfigure}
    \vspace{0.3cm}
    \caption{Ablation study results from Table \ref{tab:ablation}, illustrating the contribution of each reward function within our proposed two-stage TRIMMER framework. Performance is evaluated by comparing predicted importance scores to ground-truth annotations (averaged over 20 annotators) for the video "eQu1rNs0an0" from the TVSum dataset \cite{tvsum_dataset}.}
    \label{fig:pred_gt_eQu1rNs0an0}
\end{figure}

\begin{figure}[h]
    \centering

    \begin{subfigure}[t]{0.47\textwidth}
        \centering
        \includegraphics[width=\linewidth]{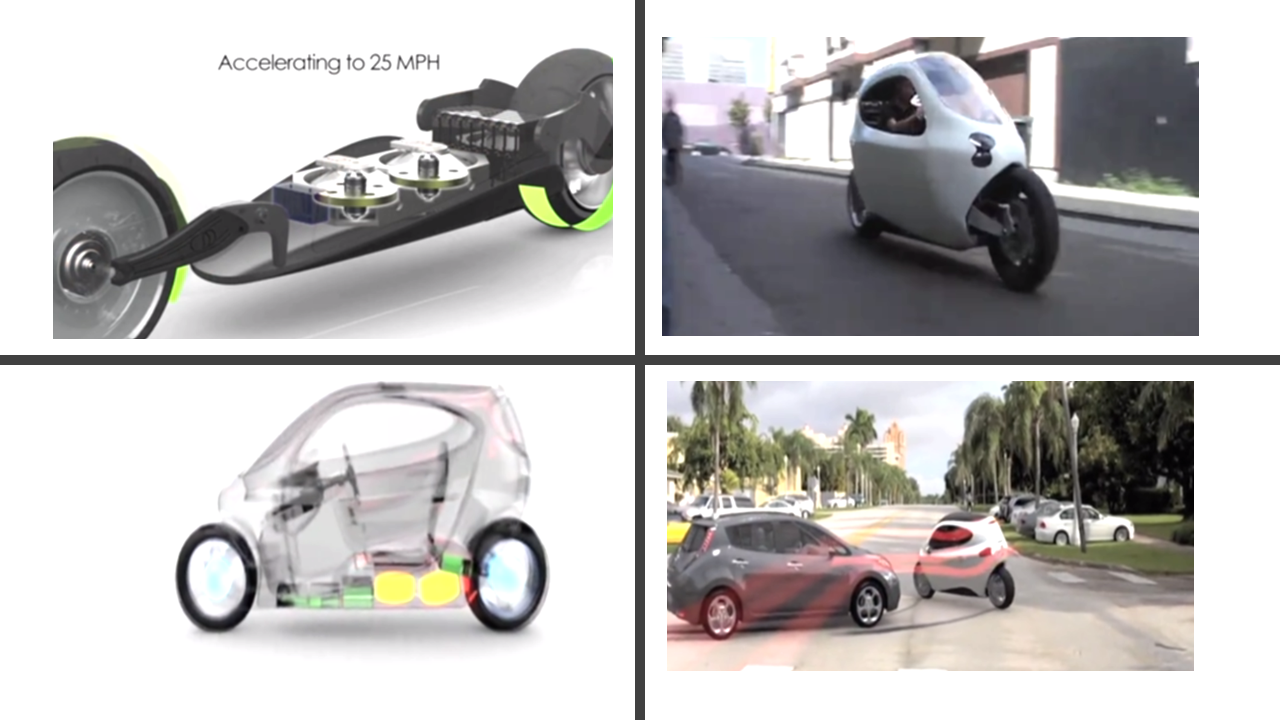}
        \caption{Key frames- Video "xwqBXPGE9pQ"}
    \end{subfigure}
    \begin{subfigure}[t]{0.5\textwidth}
        \centering
        \includegraphics[width=\linewidth]{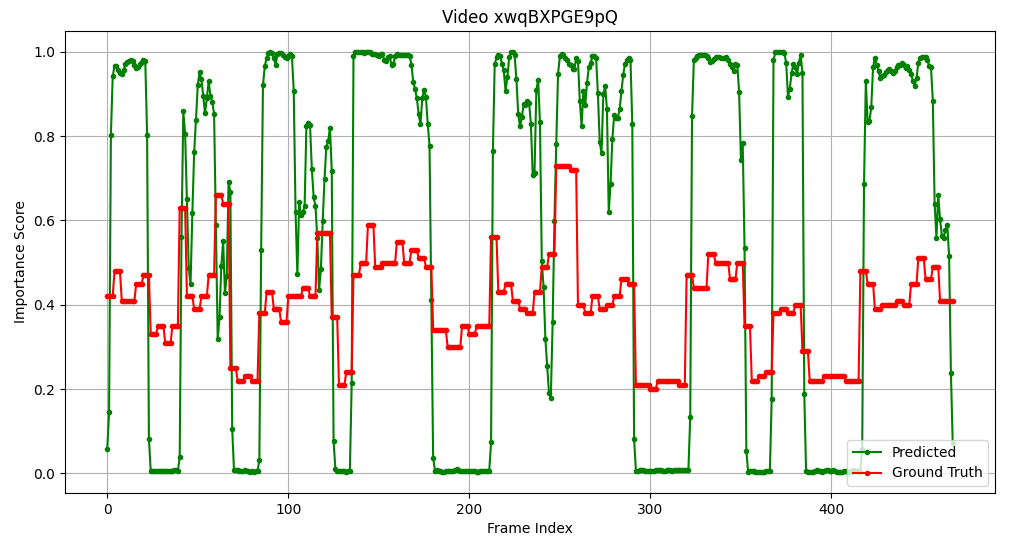}
        \caption{$R_\text{REP}$}
    \end{subfigure}

    \vspace{0.1cm}

    \begin{subfigure}[t]{0.49\textwidth}
        \centering
        \includegraphics[width=\linewidth]{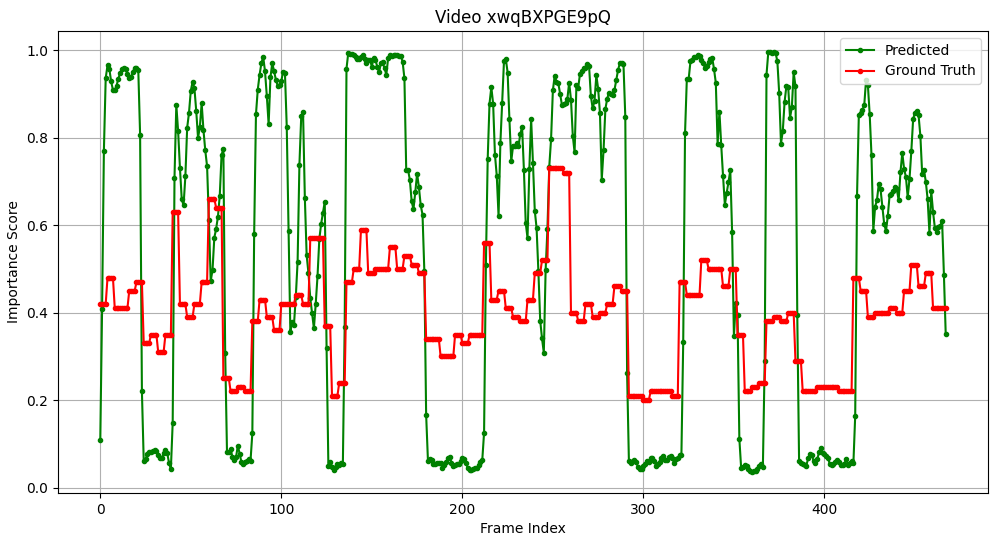}
        \caption{$R_{\mathrm{PTRIM}}$}
    \end{subfigure}
    \begin{subfigure}[t]{0.49\textwidth}
        \centering
        \includegraphics[width=\linewidth]{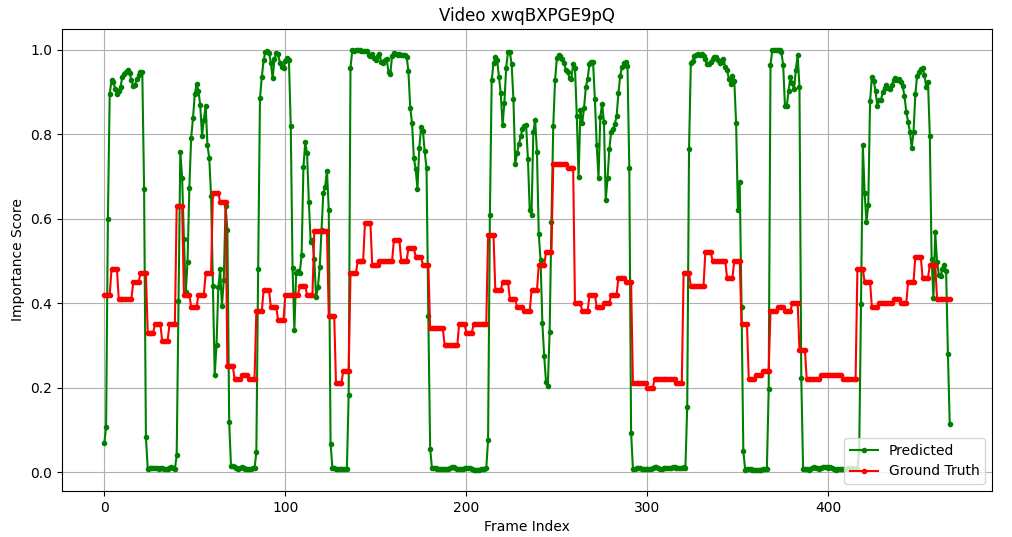}
        \caption{$R_{\mathrm{PTRIM}} + \lambda R_\text{REP}$}
    \end{subfigure}
    \vspace{0.3cm}
    \caption{Ablation study results from Table \ref{tab:ablation}, illustrating the contribution of each reward function within our two-stage TRIMMER framework. Performance is evaluated by comparing predicted importance scores to ground-truth annotations (averaged over 20 annotators) for the video "xwqBXPGE9pQ" from the TVSum dataset \cite{tvsum_dataset}.}
    \label{fig:pred_gt_xwqBXPGE9pQ}
\end{figure}

To this end, we investigate whether the empirical results support our initial hypotheses regarding the individual roles of each reward component. Figures~\ref{fig:pred_gt_eQu1rNs0an0} and~\ref{fig:pred_gt_xwqBXPGE9pQ} illustrate the behavior of the predicted importance scores generated by the cumulative reward function, as well as its individual components $R_\text{REP}$ and $R_\text{PTRIM}$. We observe a strong qualitative alignment between the predicted scores and the ground truth for both $R_\text{REP}$ and $R_\text{PTRIM}$, likely attributed to the effects of the self-supervised pretraining Stage 1. However, subtle differences between the two components emerge, which can influence their overall performance on ranked correlation with human annotations as demonstrated in Table \ref{tab:ablation}.

For instance, certain frames assigned low importance by $R_{\text{REP}}$, presumably due to limited representativeness, received higher scores under $R_{\text{PTRIM}}$, likely reflecting their contribution to diversity. In Figure~\ref{fig:pred_gt_xwqBXPGE9pQ}, from approximately frame index 400 onwards, a clear distinction between $R_\text{REP}$ and $R_\text{PTRIM}$ is observed. In this instance, neither component alone captures the ground truth pattern effectively; however, their combination through the cumulative reward function, $R = R_{REP} + \lambda R_{\mathrm{PTRIM}}$, produces a prediction that closely aligns with human-annotated importance scores.

In addition, Figure~\ref{fig:reward_ablation} shows the evolution of the reward values during training. Post pretraining, $R_\text{PTRIM}$ increases significantly, suggesting that representation learning initially emphasized representativeness over diversity. Conversely, $R_\text{REP}$ does not increase substantially, indicating that the pretraining phase has already captured representative information. When combined, the two rewards exhibit a synergistic effect: $R_\text{PTRIM}$ encourages diversity by selecting novel or complementary frames, while $R_\text{REP}$ maintains fidelity to representative content. These findings are consistent with the conceptual framework described in Section~\ref{ssec:intuition}, thereby validating our reward design through both qualitative and quantitative evidence.

\end{document}